\documentclass[runningheads]{llncs}

\usepackage{eccv}

\usepackage{eccvabbrv}
\usepackage{multirow}
\usepackage{float}
\usepackage{graphicx}
\usepackage{breqn}
\usepackage{algorithm}
\usepackage{algorithmic}
\usepackage{amsmath}
\usepackage{amssymb}
\usepackage{textcomp}
\usepackage{gensymb}
\usepackage{microtype}
\usepackage{comment}
\usepackage{lipsum}
\usepackage{pgffor}

\usepackage{graphicx}
\usepackage{booktabs}

\usepackage[accsupp]{axessibility}  
\newcommand{\myheading}[1]{\vspace{1ex}\noindent \textbf{#1}}
\RequirePackage[format=plain,labelformat=simple,labelsep=period,font=small,compatibility=false]{caption}
\def\Approach{DiverseDream}

\definecolor{mydarkblue}{rgb}{0,0.08,1}
\definecolor{mydarkgreen}{rgb}{0.02,0.6,0.02}
\definecolor{myred}{rgb}{1.0,0.0,0.0}
\definecolor{myred2}{rgb}{0.7,0.1,0.1}
\definecolor{mydarkblue2}{rgb}{0.05,0.1,0.7}
\definecolor{mypurple}{rgb}{111,0,255}
\definecolor{mypurple2}{rgb}{111,0,111}

\usepackage{hyperref}

\usepackage{orcidlink}

\begin{document}

\title{Diverse Text-to-3D Synthesis with Augmented Text Embedding} 


\author{Uy Dieu Tran$^*$ \inst{1}\orcidlink{0009-0000-4950-5235} \and
Minh Luu$^*$ \inst{1}\orcidlink{0009-0009-1756-2054} \and
Phong Ha Nguyen \inst{1}\orcidlink{0000-0002-9678-0886} \and
Khoi Nguyen\inst{1}\orcidlink{0000-0002-9259-420X} \and
Binh-Son Hua\inst{1,2}\orcidlink{0000-0002-5706-8634}}

\authorrunning{Uy.~Tran, Minh.~Luu et al.}

\institute{$^1$VinAI Research \qquad
$^2$Trinity College Dublin}
\maketitle
\def\thefootnote{*}\footnotetext{These authors contributed equally to this work}\def\thefootnote{\arabic{footnote}}

\begin{abstract}
\label{sec:abstract}
  Text-to-3D synthesis has recently emerged as a new approach to sampling 3D models by adopting pretrained text-to-image models as guiding visual priors. An intriguing but underexplored problem with existing text-to-3D methods is that 3D models obtained from the sampling-by-optimization procedure tend to have mode collapses, and hence poor diversity in their results. In this paper, we provide an analysis and identify potential causes of such a limited diversity, which motivates us to devise a new method that considers the joint generation of different 3D models from the same text prompt. We propose to use augmented text prompts via textual inversion of reference images to diversify the joint generation. We show that our method leads to improved diversity in text-to-3D synthesis qualitatively and quantitatively. Project page: \url{https://diversedream.github.io/}
\end{abstract}
\section{Introduction}
\label{sec:introduction}
The realm of 3D content creation has persistently posed intricate challenges within the domains of computer vision and computer graphics. Over time, various methodologies have emerged to address this challenge. Traditional techniques in generating 3D models often necessitate user interaction, involving meticulous shaping of scene geometry and appearance through software like Blender~\cite{blender}. Another prevalent avenue revolves around scene reconstruction using multi-view geometry principles, extensively explored in literature such as~\cite{zero123}. These approaches have garnered substantial adoption, particularly within industries like interior design and computer animation, revolutionizing their workflows and creative possibilities.

\begin{figure*}[!htp]
  \centering
  \includegraphics[width=\linewidth]{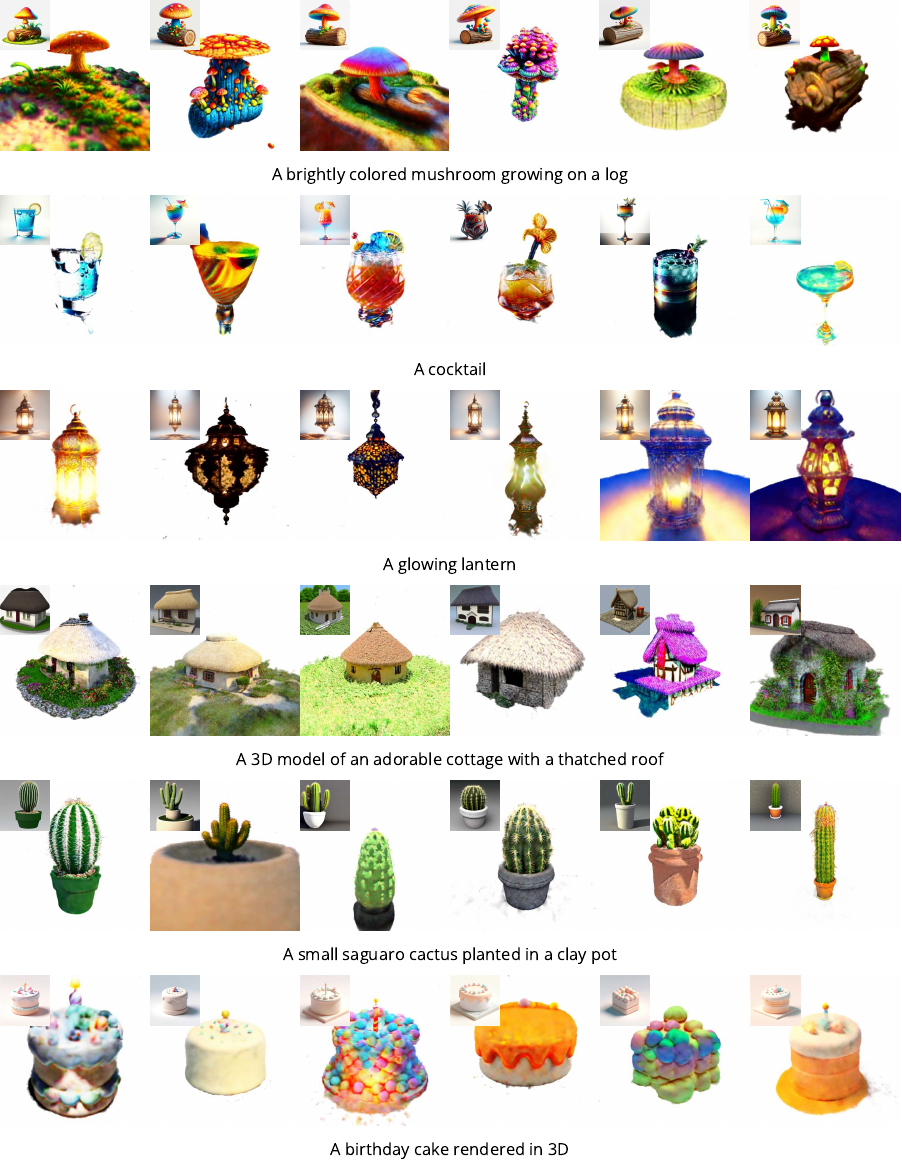}
   \caption{We address the intriguing low-diversity issue in text-to-3D synthesis by reconsidering the text prompt used by variational score distillation~\cite{wang2023prolificdreamer}. We propose to use reference images to sample augmented text prompts via textual inversion and use these augmented text prompts to condition the particles in the variational inference of text-to-3D optimization to learn more diverse 3D representations. Thanks to the diversity in the reference images (top-left inline images), we obtain diverse 3D models that inherit certain structures from their references.}
   \label{fig:diversity_result}
\end{figure*}


The rise of deep learning has led to increased interest in developing data-driven techniques to automate 3D modeling. Several efforts have been made to generate 3D models by learning directly from 3D data~\cite{shi2023survey}. However, due to the scarce availability of 3D data, it has been of great interest to explore the generation of 3D data by learning from different modalities such as images and natural languages. It has been shown that pretrained text-to-image diffusion models can serve as a strong prior to guiding the optimization of a 3D model represented by a neural radiance field in DreamFusion with SDS loss \cite{poole2023dreamfusion}, from which text-to-3D synthesis emerges as a promising research direction.

While text-to-3D synthesis has shown promises, challenges persist in fidelity, diversity, convergence, and scalability of generated models. Efforts to address these issues include enhanced loss functions like ProlificDreamer~\cite{wang2023prolificdreamer}, generalized across prompts in ATT3D \cite{lorraine2023att3d}, ATOM \cite{qian2024atom}, ET3D \cite{chen2023et3d}, and personalized generation in DreamBooth3D \cite{raj2023dreambooth3d}. Diversity, however, remains largely unexplored in current text-to-3D methods, with limited insight into its mechanisms.


In this paper, we explore methods to enhance the diversity of 3D model generation in text-to-3D systems. We posit that the diversity of model outputs is influenced by the objective function used, such as SDS~\cite{poole2023dreamfusion} and VSD~\cite{wang2023prolificdreamer}, when conditioning 3D model generation on a text prompt. Motivated by this insight, we propose a method to diversify text-to-3D generation results by augmenting the original text prompt through textual inversion techniques \cite{text_inverse, han2023hiper}. Our approach involves sampling reference images from a pretrained text-to-image diffusion model and extracting the corresponding text features via Textual Inversion. These text features are then combined with the features of the original text prompts to guide the optimization process for sampling 3D models. Experimental results (\cref{fig:diversity_result}) demonstrate a significant improvement in the diversity of generated 3D models compared to state-of-the-art methods, both quantitatively and qualitatively.

\vspace{5pt}
In summary, our contributions are:
\vspace{-5pt}
\begin{itemize}
    \item An empirical analysis of the  diversity of existing text-to-3D methods;
    \item A general technique based on augmented text embedding acquired from textual inversion of 2D reference images to improve the diversity and speed of the optimization process; 
    \item Extensive experiments and ablation studies to demonstrate the validity and robustness of our method, which is applicable to different text-to-3D methods.
\end{itemize}


\section{Related Work}
\label{sec:related_work}

\myheading{Text-to-image synthesis} has seen significant advancements, with methods relying on Generative Adversarial Networks (GANs) \cite{stylegan-t,kang2023gigagan,brock2018large} and auto-regressive models like DALL-E \cite{dall-e}, Parti \cite{yu2022scaling}, and MUSE \cite{muse}. While GANs offer fast and realistic image generation, they are prone to mode collapse. Recently, diffusion models such as Stable Diffusion \cite{stable-diffusion}, DALL-E 3 \cite{dalle-3}, and Imagen\cite{saharia2022photorealistic} have shown promise in synthesizing high-quality images. In this study, we utilize diffusion-based models, particularly Stable Diffusion, as the pretrained 2D prior to supervise our 3D model generation.

\myheading{3D representation} serves as the foundation for various 3D tasks like novel view synthesis and content creation. Neural Radiance Fields (NeRFs)~\cite{mildenhall2020nerf} have gained traction for their volumetric rendering approach, learning 3D scenes from 2D images alone. Despite NeRF's widespread use~\cite{advanceNeuralRendering}, its optimization process is time-intensive~\cite{garbin2021fastnerf}. To address this, researchers have explored hybrid scene representations like voxel grids~\cite{sun2022direct,yu2021plenoctrees}, hash-grids~\cite{muller2022instant}, tri-planes~\cite{eg3d,Chen2022ECCV}, and Gaussian splatting~\cite{kerbl3Dgaussians}, aiming to improve speed and view synthesis performance. Among these, hash-grids~\cite{muller2022instant} are favored for text-to-3D tasks~\cite{poole2023dreamfusion,wang2023prolificdreamer} due to their fast training and superior performance compared to NeRFs. In this paper, we leverage hash grids to learn diverse 3D scene representations from a single text prompt using our proposed textual score distillation loss.

\myheading{Image-to-3D generation} is a crucial aspect of conditional generative 3D models. Early methods like SynSin \cite{wiles2020synsin} and Free View Synthesis \cite{nguyen2022free} rely on differentiable neural renderers for single view synthesis but are limited by pose distances and struggle with full 360\degree reconstructions from a single input. Recent advancements have seen models like Zero-1-to-3 \cite{zero123} pioneering open-world single-image-to-3D conversion through zero-shot novel view synthesis, yet face challenges with geometric consistency. Works such as One-2-3-45 \cite{One-2-3-45}, SyncDreamer \cite{SyncDreamer}, LRM \cite{hong2024lrm}, LGM \cite{tang2024lgm}, and Consistent123 \cite{consistent123} address this by adding geometry-constraint layers to improve consistency. However, these methods typically require extensive 3D model datasets like ShapeNet \cite{chang2015shapenet} or large-scale multiview datasets like Objaverse \cite{objaverse} for training. In contrast, our approach solely relies on a pretrained text-to-image model for supervision, making it more accessible.

\myheading{Text-to-3D generation} has seen remarkable progress recently, leveraging pre-trained text-to-image models like Stable Diffusion \cite{stable-diffusion}. Early works like DreamField \cite{dreamfields} use CLIP \cite{clip} to align rendered images with input text but often compromise on model quality due to CLIP's limited semantic feature capture. DreamFusion \cite{poole2023dreamfusion} substitutes CLIP loss with Score Distillation Sampling (SDS) and introduces efficient gradient calculation for neural radiance field learning. However, it tends to produce oversmooth surfaces and saturated colors. Subsequent methods aim to address these limitations by enhancing resolution \cite{magic3d}, appearance \cite{fantasia3d, zhu2024hifa, katzir2023noisefree, yu2024texttod}, geometry \cite{shi2024mvdream, li2024instantd, anonymous2024dreamcontrol, qian2024magic, seo2024let}, speed \cite{dreamgaussian, huang2024dreamtime, li2024instantd}, and photorealism \cite{katzir2023noisefree, wang2023prolificdreamer,lee2024dreamflow, zhu2024hifa}. Despite this progress, diversity in text-to-3D synthesis remains underexplored, motivating our work.


\myheading{Textual inversion}. While recent text-to-image diffusion models like Stable Diffusion \cite{stable-diffusion} and DALL-E 3 \cite{dalle-3} produce high-quality 2D images, they may not preserve the subject's shape or identity, known as ``personalization''. Techniques such as Textual Inversion \cite{text_inverse} and DreamBooth \cite{ruiz2022dreambooth} aim to maintain subject identity in reference images by introducing a virtual token whose embedding can be optimized to manipulate the generated images. HiPer inversion \cite{han2023hiper}, building upon Textual Inversion, enhances inversion by using a single reference image and optimizing textual tokens in the text prompt to store object identity. Inspired by this, we apply textual inversion to diversify generated 3D content.

\section{Background}
\label{sec:analysis}

A typical approach to text-to-3D synthesis is to leverage the 2D prior from a pretrained text-to-image model such as Stable Diffusion (SD) \cite{stable-diffusion}, to guide the training of a 3D model represented by a neural radiance field (NeRF). In particular, a NeRF parameterized by $\theta$ is optimized so that its rendered images $x=g(\theta, c)$, with $g$ as the volumetric rendering function and $c$ as the camera pose, look realistic and conform to the text prompt $y$. 

\myheading{Score distillation sampling (SDS):}  DreamFusion \cite{poole2023dreamfusion} introduced the SDS loss whose gradient is computed as:
\begin{align} \label{eq:sds}
    \nabla_\theta \mathcal{L}_{\text{SDS}} \triangleq \mathbb{E}_{t, \epsilon,c} \left[\omega(t)(\epsilon_{\text{SD}} (x_t, t, y)-\epsilon) \frac{\partial g(\theta, c)}{\partial \theta}\right],
\end{align}
where $\omega(t)$ is a time-dependent weighting function, $\epsilon_{\text{SD}}$ is the predicted noise of SD given the noisy input image $x_t=\alpha_t x + \sigma_t\epsilon$ created by adding Gaussian noise $\epsilon$ to the rendered image $x$ at timestep $t$ with noise scheduling coefficients $\alpha_t, \sigma_t$. However, the SDS loss often suffers from over-saturation, over-smoothing, and low-diversity issues as empirically analyzed in \cite{wang2023prolificdreamer}.
The low diversity issue in SDS becomes apparent when multiple runs yield similar results empirically.
Therefore, we advocate the use of the more sophisticated variational score distillation loss \cite{wang2023prolificdreamer} for our exploration of the diversity of text-to-3D synthesis, which we briefly describe below.

\myheading{Variational score distillation (VSD):} ProlificDreamer \cite{wang2023prolificdreamer} mitigates the limitations of DreamFusion by introducing a variational form of score distillation. Their VSD loss aims to tackle the low-diversity issue by modeling the distribution $\mu$ of 3D models $\theta$ generated from a single text prompt $y$ as $\mu(\theta | y)$. It's worth noting that SDS is a special case of VSD, where $\mu(\theta | y)$ simplifies to a Dirac distribution $\delta(\theta - \theta^1)$, resulting in only a single 3D model $\theta^1$ for each text prompt.

To optimize VSD, the distribution $\mu$ is approximated by $K$ learnable particles where each particle $i$ corresponds to a 3D representation parameterized by $\theta_i$ which is sampled from a set of $K$ particles $\{\theta_i\}_{i=1}^K$ for each training iteration, following the particle-based variational inference framework.
The gradient of the VSD loss is as follows:
\begin{align} \label{eq:vsd}
    \nabla_{\theta_i} \mathcal{L}_{\text{VSD}} \triangleq \mathbb{E}_{t, \epsilon,c} \left[ \omega(t)(\epsilon_{\text{SD}} (x_t^i, t, y) - 
     \epsilon_\phi(x_t^i, t, c, y)) 
     \frac{\partial g(\theta_i,c)}{\partial \theta_i} \right],
\end{align}
where $\epsilon_\phi$ is a fine-tuned version of the original SD using the LoRA \cite{hu2021lora} parameterization $\phi$ on the rendered images of in-progress learning NeRFs. LoRA can be regarded as the \emph{domain adaptation} of SD to noisy images rendered from NeRFs since SD is not originally trained on noisy images. Specifically, $\phi$ is trained with the following objective: 
\begin{align} \label{eq:phi}
    \min_\phi  \mathbb{E}_{t, \epsilon,c} \left[ \| \omega(t)(\epsilon_{\phi} (x_t^i, t, c, y)-\epsilon) \|^2_2 \right].
\end{align}
Beyond its theoretical modeling of 3D representations as a distribution, an empirical observation on why the VSD loss enhances the diversity compared to the SDS loss is that the objective of VSD for each particle is different from each other. Notably, the second term $\epsilon_{\phi}(x_t^i, t, c, y)$ (in Eq.~\eqref{eq:vsd}) dynamically changes due to the learning progression of $\phi$ and the input image $x_t^i$ rendered from the current particle $\theta_i$. 

Although the VSD loss addressed the limitation of SDS loss and clearly improved the quality of the 3D representations, we empirically found that it still yields limited diversity in some particular prompts.
To further improve diversity, we propose to use augmented text embedding guided by 2D reference images, which is presented in the next section.

\section{Our Approach} 
\label{sec:approach}

\subsection{Analysis} 

\begin{figure*}[t]
  \centering
  \includegraphics[width=.8\linewidth]{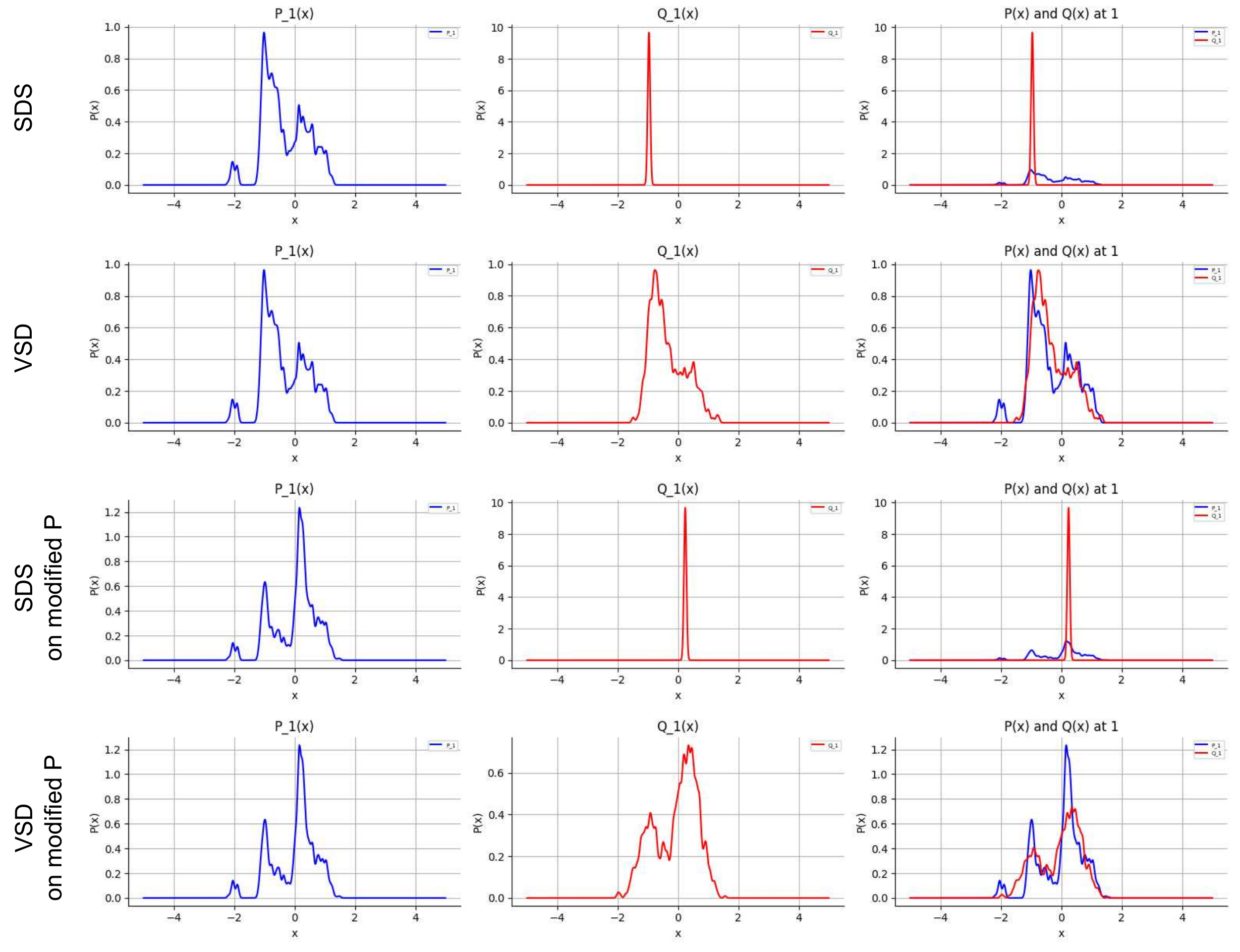}
   \caption{
    We present a simulation of SDS (first row) and VSD (second row) in KL form on a 1D toy dataset, where the ground truth distribution $p_{\text{SD}}( x_{t}|y)$ is a 7-component Gaussian mixture model. Results are shown at $t=1$ (low noise data). In the third row and forth row, varying $p_{\text{SD}}( x_{t}|y')$ with a new text prompt $y'$ leads to diverse outcomes across different runs with SDS/VSD loss, motivating our approach.
   }\label{fig:toy_data}
   \vspace{-10pt}
\end{figure*}

Let us first motivate our method by an empirical analysis on the diversity of SDS and VSD loss, the prevailing loss functions for generating 3D assets from a given text prompt using a text-to-image model as prior. 
A notable drawback of this technique is that the SDS loss often yields almost identical results across different runs, primarily due to the mode-seeking behavior exhibited by the KL divergence between a Gaussian distribution and a multi-modal landscape of the text-to-image prior. More precisely, as shown from \cite{poole2023dreamfusion}, the SDS loss in the KL form is given by:
\begin{align} \label{eq:sds_kl}
   \mathcal{L}_{\text{SDS}} \triangleq \mathbb{E}_{t}\left[\text{KL}(q(x_{t} |x=g( \theta,c )) ||p_{\text{SD}} ( x_ {t}| y))\right].
\end{align}
At a given time step $t$, $q(x_{t}|x)=\mathcal{N}( \alpha_{t} x, \sigma _{t}^ {2} \mathbf{I})$ represents a Gaussian distribution characterizing the forward diffuse process of the rendered image, while $p_{\text{SD}}(x_{t}|y) = \int p_{\text{SD}}^0 ( x_{0} |y) q( x_{t}|x_0)dx_0$ denotes the marginal distribution of the diffusion model. It is reasonable to assume that the distribution of the diffusion model exhibits multimodality, particularly for lower values of $t$. Considering that $q(x_{t}|x)$ is a unimodal distribution, it tends to align with the closest mode of $p_{\text{SD}}(x_{t}|y)$, as demonstrated by \cite{bishop2006PRML_inbook}. This behavior is particularly pronounced in 3D, as it necessitates initializing the 3D scene as a Gaussian blob in each run. Consequently, the initial parameters of NeRF $\theta$ tend to be closer to each other in successive runs, resulting in low-diversity outcomes for SDS.

Meanwhile, the KL form of VSD (\cref{eq:vsd_kl}) is nearly identical to SDS, albeit the substitution of the unimodal Gaussian $q(x_{t}|x)$ with a more intricate, implicit distribution $q^{\mu}( x_{t} |c, y)=\mathbb{E}_{\mu (\theta |y)}[q( x_ {t} |x=g( \theta,c ))]$.
\begin{align} \label{eq:vsd_kl}
   \mathcal{L}_{\text{vsd}} \triangleq \mathbb{E}_{t}\left[\text{KL}(q^{\mu}( x_{t} |c, y) ||p_ {\text{SD}} ( x_ {t}|y))\right]
\end{align}
By increasing the complexity of $q^{\mu}$, the optimized distribution $q^{\mu*}$ will possess a greater capacity to accurately fit the target distribution $p_{SD }( x_ {t}|y)$. The VSD algorithm aims to draw data from $\mu^{*} (\theta |y)$ to minimize \cref{eq:vsd_kl} through particle-based variational inference. As a result, the final 3D assets will exhibit a greater degree of variety. 

We elucidate our intuition in \cref{fig:toy_data} by performing a simulation for SDS and VSD on a 1D toy dataset, where $p_ {\text{SD}} ( x_ {t}|y)$ is a Gaussian mixture model (GMM) comprising 7 components. We select a random position $x$ as the optimizable parameter for SDS, while the means and standard deviations of a 3-component GMM $\mu (\theta |y)$ serve as the optimizable parameters for VSD. 
Based on this analysis, instead of modeling $q(x_t | x)$ as done in SDS and VSD, we propose a new way to increase the diversity of the text-to-3D models by modifying the distribution $p_{\text{SD}}(x_t | y)$. Our idea is to \textbf{diversify the condition $y$ to $y'$} so that optimization with the prior $p_{\text{SD}} ( x_{t}|y')$ potentially yields a more substantial impact due to the alteration in the optimization landscape.

\subsection{Method Overview} 

To implement the idea of diversifying $y$, inspired by textual inversion methods~\cite{text_inverse,han2023hiper} for 2D object personalization, we aim to condition text-to-3D synthesis such that the per-particle difference in $\epsilon_{\text{SD}}(x_t^i, t, c, y'_i)$ is boosted by using different and distinct prompt $y'_i$ for each particle $\theta_i$. To this end, we devise a new approach that leverages HiPer textual inversion \cite{han2023hiper} to enhance the resulting diversity. Here we base our discussion on VSD, but the idea generalizes to SDS as well.



\begin{figure*}[t]
  \centering
  \includegraphics[width=1\linewidth]{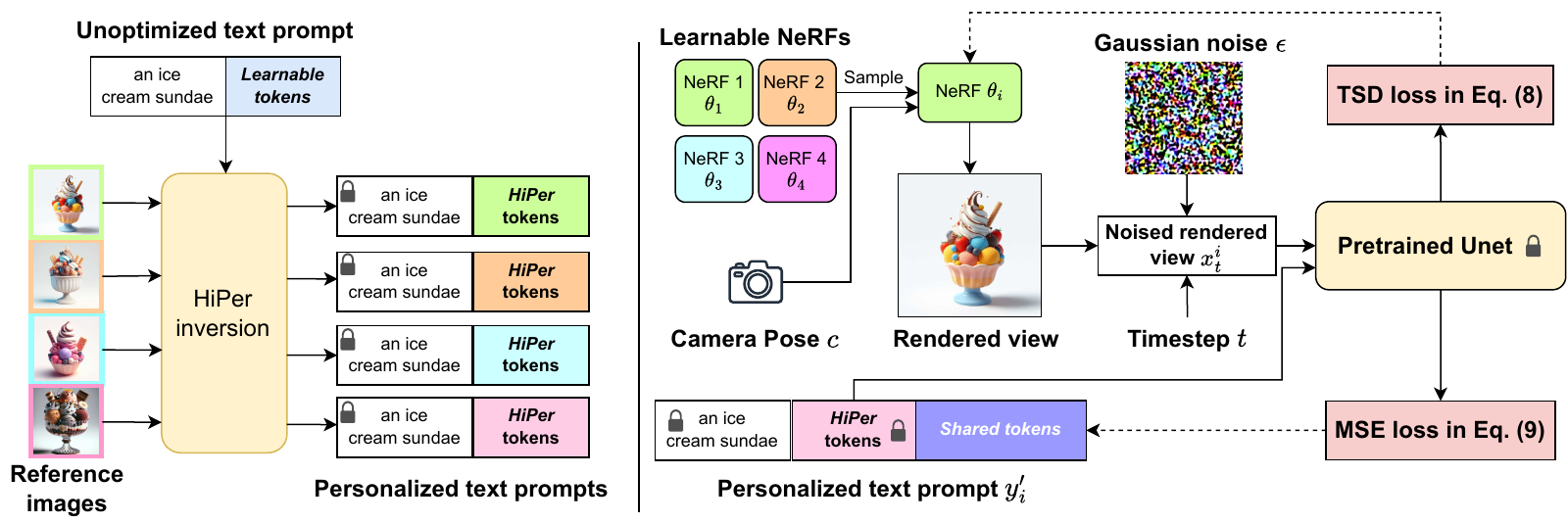}
   \caption{We translate the diversity of augmented text prompts to the resulting 3D models via a two-stage method. \textbf{Stage 1: HiPer tokens inversion} (left): for each reference image, we seek to learn a HiPer token $h_i$ so that the prompt $[y; h_i]$ reconstructs the reference image.
   \textbf{Stage 2: Textual score distillation} (right): we run a multi-particle variational inference for optimizing the 3D models from text prompt $y$. For each iteration in the optimization, we randomly sample a particle $\theta_i$ with its rendered image $x_i$. We use the augmented text prompt $y'_i = [y; h^*_i;\phi]$, with $\phi$ as shared embedding to condition the optimization of $\theta_i$ (Eq.~\eqref{eq:tsd} and Eq.~\eqref{eq:textual_phi}). 
   }\label{fig:diagram}
   \vspace{-10pt}
\end{figure*}

Our approach consists of two stages: HiPer tokens inversion and textual score distillation. Firstly, we select a reference image $x^r_i$ for each particle and determine an optimized HiPer token $h^*_i$. This token is chosen such that the reference image can be faithfully reconstructed by the pretrained text-to-image diffusion model with the prompt $[y; h^*_i]$, where $[;]$ denotes concatenation. In the second stage, multiple particles $\theta_i$ are collectively optimized alongside a new shared domain adapter $\phi$, which is also encoded as a learnable token. This process forms the augmented text prompt $[y; h^*_i; \phi]$ for each particle. The algorithm is depicted in Alg.~\ref{alg:tsd} and illustrated in \cref{fig:diagram}.


\begin{algorithm}[th]
\footnotesize
\caption{Algorithm of \Approach.}
\label{alg:tsd}
\textbf{Input: }{$K$ particles, $K$ reference images $\{x^r_i\}_{i=1}^K$ from prompt $y$, pretrained text-to-image model $\epsilon_{\text{SD}}$.}

\textbf{Stage 1: HiPer tokens inversion}

\begin{algorithmic}[1]
\STATE {\bfseries initialize} $K$ HiPer tokens $\{h_{i}\}_{i=1}^{K}$.
\FOR{i=1 \TO K} 
    \STATE Optimize $h_i$ given $x^r_i$ following ~\cref{eq:hiper} to obtain $h_i^*$.
\ENDFOR
\STATE {\bfseries return} optimized $\{h_i^*\}_{i=1}^K$
\end{algorithmic}

\textbf{Stage 2: Textual score distillation} 

\begin{algorithmic}[1]
\STATE {\bfseries initialize} $K$ NeRFs $\{\theta_{i}\}_{i=1}^{K}$, shared learnable tokens $\phi$.

\WHILE{not converged}
\STATE Sample noise $\epsilon$, camera pose $c$, timestep $t$, and index $i$, obtain $\theta_i$, and form text prompt $y'_i = [y;h_i^*;\phi]$.
\STATE Render image $x_i=g(\theta_i,c)$ from NeRF $\theta_i$ at pose $c$, and compute $x^i_t$.
\STATE Update $\theta_i$ following \cref{eq:tsd}.
\STATE Update $\phi$ following \cref{eq:textual_phi}.
\ENDWHILE
\STATE {\bfseries return} optimized $\{\theta_i^*\}_{i=1}^K$
\end{algorithmic}
\end{algorithm}

\subsection{HiPer tokens inversion} 
We first sample $K$ reference images  $\{x_i^r\}_{i=1}^K$ corresponding to $K$ particles from any text-to-image model given text prompt $y$. We empirically find that using Stable Diffusion (SD) \cite{stable-diffusion} with additional guidance like ``X with white background'' gives the most suitable images for HiPer textual inversion. This is because we only have one image for inversion and we want to exclude noisy factors like background, facilitating faster and better textual inversion. 

Subsequently, we want to optimize HiPer tokens for each reference image using the technique in \cite{han2023hiper}. Specifically, given a reference image $x^r_i$ and a text prompt $y \in \mathbb{R}^{L_1 \times D}$ with $L_1$ as the number of text tokens and $D$ as feature dimensions, we seek to find HiPer tokens $h \in \mathbb{R}^{L_2 \times D}$ with $L_2$ as the number of HiPer tokens to reflects the personalized identity of the object in $x^r_i$.  
To this end, we append the learnable tokens $h_i$ to the original text prompt $y$ to form new text personalized text prompt $y_i=[y; h_i] \in \mathbb{R}^{(L_1 + L_2) \times D}$, and use HiPer \cite{han2023hiper} to optimize $h_i$ with the objective: 
\begin{align} \label{eq:hiper}
    \min_{h_i}  \mathbb{E}_{t, \epsilon} \left[ \|\omega(t)\epsilon_{\text{SD}} (x_{t,i}^{r}, t, [y; h_i])-\epsilon \|^2_2 \right].
\end{align}
Note that HiPer use Stable Diffusion (version 1.4) for textual inversion. This stage is visualized in Fig.~\ref{fig:diagram} (Left).
The optimized $h_i^*$ is leveraged as the key component to diversify the results of text-to-3D synthesis in the next step.


\begin{figure}[t]
  \centering
  \includegraphics[width=0.6\linewidth]{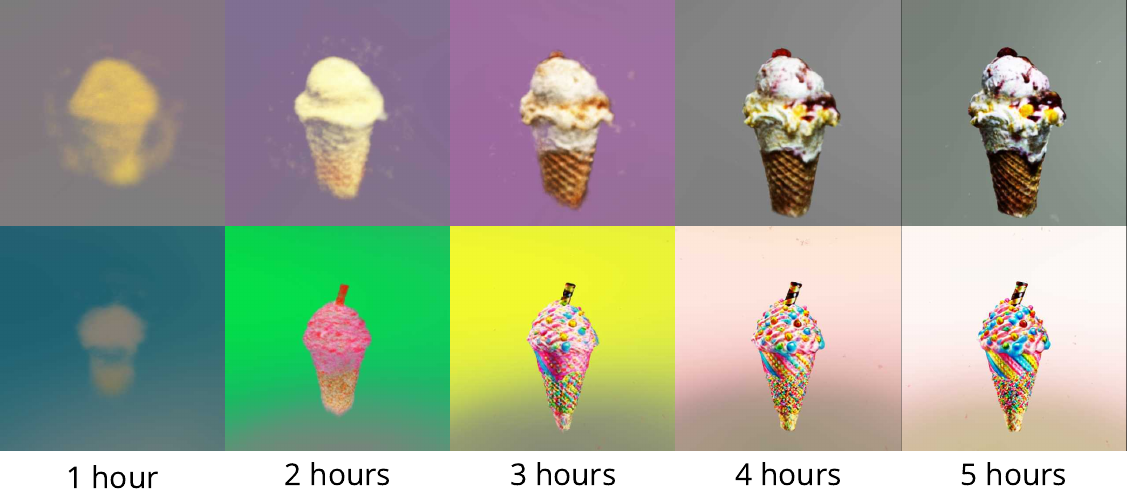}
  \vspace{-10pt}
  \caption{Optimization progress of VSD (upper) vs ours (lower). TSD with less \#learnable parameters converges faster than VSD. Prompt: ``A high-quality ice cream sundae''.
   }\label{fig:progression}
   \vspace{-10pt}
\end{figure}

\subsection{Textual score distillation (TSD)}

With the learned personalized text prompts $y_i = [y; h^*_i]$, we are ready to use them to replace the original text prompt $y$ in any text-to-3D approaches such as ProlificDreamer \cite{wang2023prolificdreamer} to enhance the diversity of these approaches. However, we discover that the Domain Adaptor $\phi$ in ProlificDreamer, which is implemented using LoRA \cite{hu2021lora}, can be further replaced by the textual inversion technique like HiPer \cite{han2023hiper}. This is similar to the problem of 2D object personalization where LoRA Dreambooth \cite{ruiz2022dreambooth} can be replaced by Textual Inversion \cite{text_inverse} or HiPer \cite{han2023hiper} with similar performance. The observation motivates us to devise a new Domain Adaptor $\phi \in \mathbb{R}^{L_3 \times D}$ in the form of shared learnable tokens in the text prompt among particles. That is, the new personalized text prompt:
\begin{align}
    y'_i = [y; h^*_i; \phi] \in \mathbb{R}^{(L_1 + L_2 + L_3) \times D},
\end{align}
where $L_3$ is the number of shared learnable tokens. The new text prompt $y'_i$ can replace the LoRA implementation $\phi$ of ProlificDreamer, resulting in the following Textual Score Distillation (TSD): 
\begin{align} \label{eq:tsd}
    &\nabla_{\theta_i} \mathcal{L}_{\text{TSD}} \triangleq \mathbb{E}_{t, \epsilon,c} \left[ \omega(t)(\epsilon_{\text{SD}} (x_t^i, t, y_i) - 
     \epsilon_{\text{SD}}(x_t^i, t, y'_i)) 
     \frac{\partial g(\theta_i,c)}{\partial \theta_i} \right].
\end{align}
Compared to the LoRA implementation, the term $\epsilon_{\text{SD}}(x_t^i, t, y'_i)$ with shared learnable tokens has the advantage of faster training speed since the number of our learnable parameters $\phi$ (about 30K parameters) is much smaller than those of LoRA (about 1.3M parameters).
 The training of the shared learnable tokens is similar to the LoRA implementation, i.e., via a separate updating step from the updating step of each particle as: 
\begin{align} \label{eq:textual_phi}
    \min_\phi  \mathbb{E}_{t, \epsilon,c} \left[ \| \omega(t)(\epsilon_{\text{SD}} (x_t, t, y'_i)-\epsilon) \|^2_2 \right].
\end{align}
In Fig.~\ref{fig:diagram} (Right), we show how we train the sampled NeRF model $\theta_i$ and shared token $\phi$ using the proposed TSD (\cref{eq:tsd}) and MSE (\cref{eq:textual_phi}) losses respectively.
As can be seen in the Fig.~\ref{fig:progression}, our method can produce higher quality samples than those produced by VSD~\cite{wang2023prolificdreamer} given the same amount of optimization time.



\section{Experiments}
\label{sec:experiments}

\myheading{Metrics.}  The common metrics for generative models such as FID \cite{NIPS2017_8a1d6947} do not separately measure fidelity and diversity. 

Inspired by \cite{anonymous2023taming}, we proposed to use a modified version of \emph{Inception Quality (IQ)} and \emph{Inception Variance (IV)} to measure the quality and diversity of our models. Our IQ and IV are formulated as follows:
\begin{align}
\text{IQ}(\theta)=\mathbb{E}_{i, c} \left[ \mathcal{H}[p(y \mid x_i= g(\theta_i, c))] \right], \\
\text{IV}(\theta)= \mathcal{H} \left[ \mathbb{E}_{i,c}[p(y \mid x_i = g(\theta_i, c))] \right],
\end{align}
where $p(y \mid x_i = g(\theta_i, c))$ is the pretrained classifier given the rendered images $x_i$ from particles $i$. The entropy $\mathcal{H}$ serves as an indicator of the classifier's confidence when presented with an input-rendered image. The IQ metric captures the expected entropy, reflecting the classifier's certainty across all views, which indicates the image quality to some degree \textit{(the lower the better)}. Conversely, the IV metric quantifies the entropy of the expected classifier outputs \textit{(the higher the better)}. A higher IV score is achieved when the classifier outputs are uniformly distributed, indicating greater diversity in the input images. 
We rendered 120 views for each particle to compute IQ and IV.

We also propose the \emph{Cosine Sim} metric to quantify the diversity of our particles. Specifically, for a given set of $K$ particles, we render the same view for each particle. The rendered images are then fed through a feature extractor (e.g., as DINO \cite{zhang2022dino}) to obtain feature vectors. We then calculate the cosine similarity for these feature vectors across $\binom{K}{2}$ pairs and take the average. To ensure robustness, we also average the results over 120 rendered views.

\myheading{Implementation details.}
Our method is implemented with threestudio, an open-source framework for text-to-3D synthesis~\cite{threestudio2023}. We use $K=6$ particles for both VSD and our framework, TSD. The training of all experiments is conducted for 50K iterations. We use $L_2=5$ as HiPer tokens optimized in 1.4K iterations. Also, we use $L_3=8$ as shared learnable tokens $\phi$ and trained in 50K iterations. For the camera embedding, we use the same implementation as threestudio \cite{threestudio2023}.
Regarding the resolution, we train each particle at $256 \times 256$ for all methods. 

\begin{figure}[t]
  \centering
  \includegraphics[width=.9\linewidth]{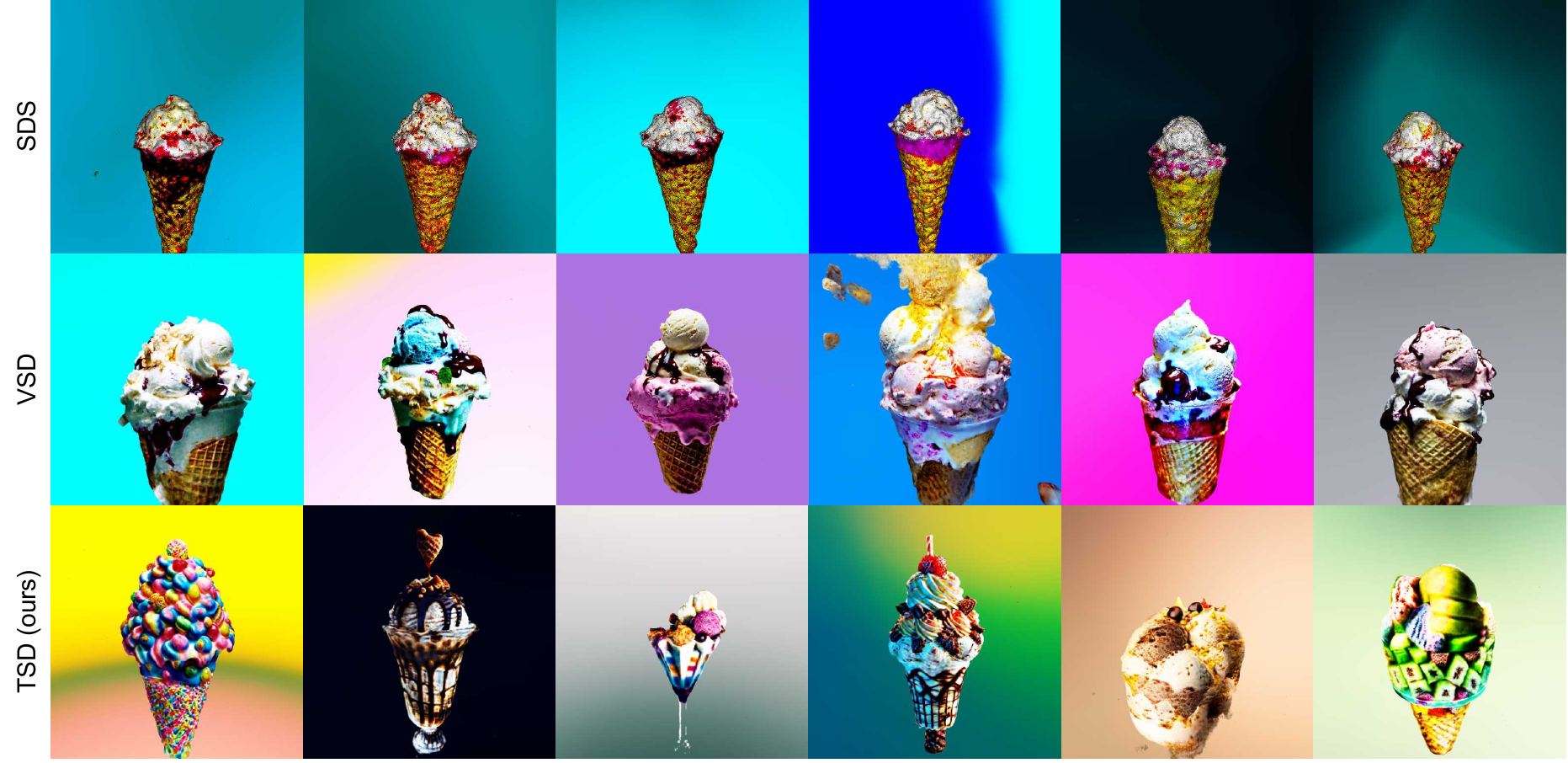}
   \caption{Diversity comparison between SOTAs and our method.}
   \label{fig:comparison}
\end{figure}

\begin{table}[htbp]
\footnotesize
\setlength{\tabcolsep}{0pt}
\begin{minipage}[t]{0.45\linewidth}
\centering
\caption{Comparison with SOTAs}
\label{tab:quant}
\begin{tabular}{lccc}
        \toprule
         & \textbf{IQ $\downarrow$} &  \textbf{IV $\uparrow$} & \textbf{Cosine Sim $\downarrow$} \\
        \midrule
        SDS~\cite{poole2023dreamfusion} & 3.695 & 4.577 & 0.720  \\
        VSD~\cite{wang2023prolificdreamer} & \textbf{3.345} & 4.586 & 0.476  \\
        \midrule
        TSD (ours) & 3.614 & \textbf{5.075} & \textbf{0.380}  \\
        \bottomrule
\end{tabular}
\end{minipage}
\begin{minipage}[t]{0.5\linewidth}
\centering
\caption{Study on \#HiPer tokens.}
\label{tab:tab3_4}
\begin{tabular}{cccc}
    \toprule
    \textbf{\#tokens} & \textbf{IQ $\downarrow$} &  \textbf{IV $\uparrow$} & \textbf{Cosine Sim $\downarrow$} \\
    \midrule
    1 & \textbf{3.375} & 4.908 & \textbf{0.403}  \\
    5 & 3.790 & 4.886 & 0.415  \\
    10 & 4.428 & \textbf{5.138} & 0.409  \\
    15 & 4.721 & 4.862 & 0.445  \\
    20 & 5.193 & 4.978 & 0.425  \\
    \bottomrule
\end{tabular}
\end{minipage}
\vspace{-20pt}
\end{table}

\subsection{Comparison with Prior Methods}

\myheading{Baselines.} We compare our method with two prominent text-to-3D methods including DreamFusion \cite{poole2023dreamfusion} and ProlificDreamer \cite{wang2023prolificdreamer}. Note that we do not compare to other variants such as Magic3D \cite{magic3d} or Fantasia3D \cite{fantasia3d} since these methods address different issues of SDS, which is orthogonal to our method which focuses on diversity. 
For validation, we select a set of 60 text prompts from DreamFusion \cite{poole2023dreamfusion} and 10 text prompts generated randomly from ChatGPT \cite{OpenAI2023GPT4TR}.

\begin{figure}[t]
  \centering
  \vspace{10pt}
  \includegraphics[width=0.85\linewidth]{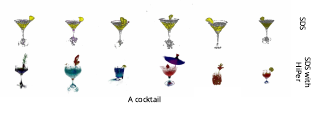}
   \vspace{-15pt}
   \caption{We demonstrate that our method remains effective for SDS.}
   \label{fig:sds_hiper}
\end{figure}

\begin{figure}[t]
  \centering
  \includegraphics[width=0.85\linewidth]{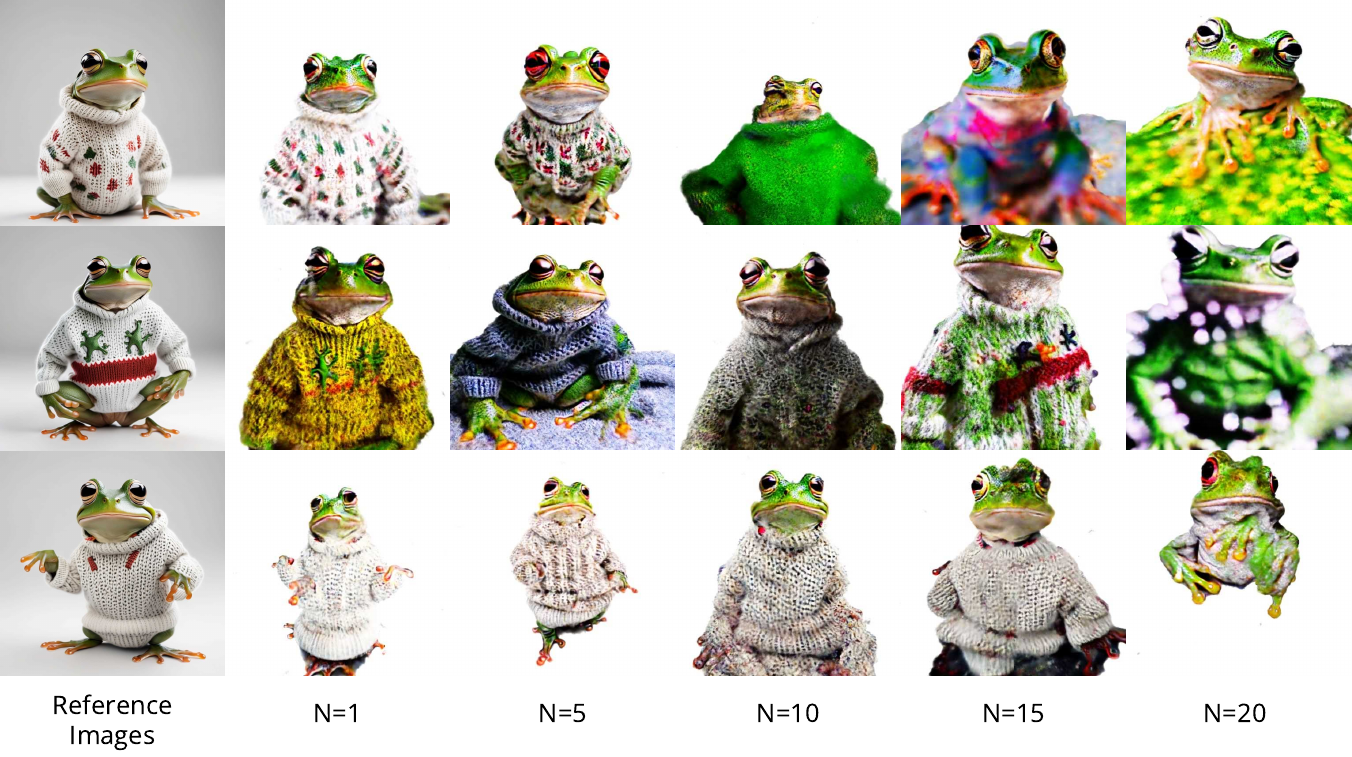}
   \caption{Visual results with different numbers of HiPer tokens.}
   \vspace{-10pt}
   \label{fig:ablation-nhiper}
\end{figure}

\begin{table}[t]
\centering
\footnotesize
\setlength{\tabcolsep}{1pt}
    \caption{Ablation study of our method.}
    \label{tab:ablation_lora}
    \begin{tabular}{llcccc}
        \toprule
        $\epsilon_{\phi}$ & \textbf{HiPer} &  \textbf{IQ} $\downarrow$ & \textbf{IV} $\uparrow$ & \textbf{Cosine Sim} $\downarrow$ & \textbf{Training time (hours)} \\
        \midrule
        LoRA & No & \textbf{3.345} & 4.586 & 0.476  & 10.23 \\
        LoRA & Yes & 3.662 & \textbf{5.109} & \textbf{0.355} & 9.50  \\
        Shared tokens & Yes & 3.614 & 5.075 & 0.380 & \textbf{7.16} \\
        \bottomrule
    \end{tabular}
\end{table}

\myheading{Quantitative comparison.} We present our quantitative results in Tab.~\ref{tab:quant}. It is evident that we outperform VSD and SDS in terms of IV Score and Cosine Sim, which measure the diversity between the particles. The IQ score of our method is slightly lower than VSD, which demonstrates there remains some fidelity-diversity trade-off, which is well known for existing text-to-image methods. 
Our method outperforms SDS in both diversity and quality.

\myheading{Qualitative comparison.}
The comparative results in Fig.~\ref{fig:comparison} demonstrate that our method offers more diversity among particles compared to VSD and SDS. For example, when given ``A high-quality ice cream sundae'' prompts, VSD tends to collapse into cone-shaped ice cream, while our method is capable of generating glass shapes and other variants.
Our 3D models inherit texture and structure from reference images (see Fig.~\ref{fig:diversity_result}), showcasing the potential of transferring 2D diversity to 3D through text prompt augmentation. Perfect inversion by HiPer is not necessary; capturing the essence of reference images suffices for diversity among personalized text prompts.
\cref{fig:sds_hiper} demonstrates the diversity of our results when applying HiPer to the SDS loss.

\begin{figure}[t]
  \centering
  \includegraphics[width=0.85\linewidth]{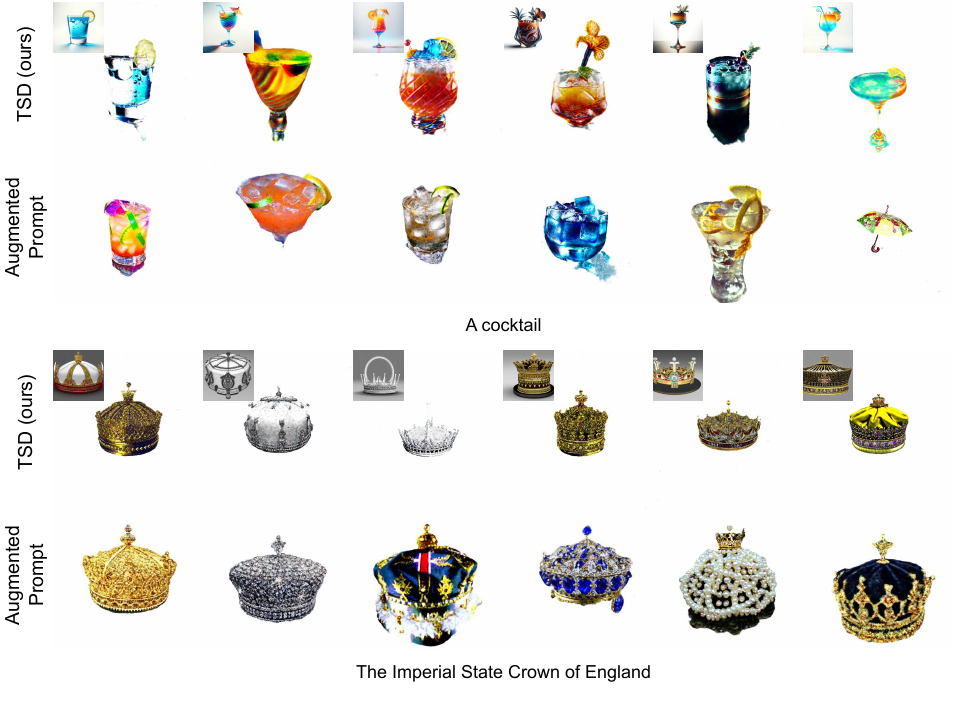}
  \vspace{-10pt}
   \caption{Comparison between our HiPer inversion with LLM-generated augmentation for text prompt sampling. It can be seen that the use of reference images in our method leads to better fidelity and diversity.}
  \vspace{-10pt}
   \label{fig:augment_text2}
\end{figure}

\subsection{Ablation Study}
In this section, we undertake an ablation study to examine the factors that influence our methods.

\myheading{Number of HiPer tokens $L_2$.}
We vary the number of the HiPer tokens from 1, 5, 10, 15, and 20 using the prompt ``A DSLR photo of a frog wearing a sweater'', Fig.~\ref{fig:ablation-nhiper} and Tab.~\ref{tab:tab3_4} show that while altering token length has a subtle impact on the 3D model, the text-to-image model produces images more closely resembling the reference in 2D.




\myheading{LoRA vs. shared learnable tokens.} We further validate the effect of our shared learnable tokens. 
When replacing the LoRA layer with our learnable tokens, our method can still achieve diverse results compared to SDS and VSD, although the quality was not as good as our method with LoRA, as shown in Tab.~\ref{tab:ablation_lora}. However, the advantage of using shared learnable tokens is that it achieves a better training speed compared to the use of LoRA. 

\myheading{LLM-based text prompt augmentation.}
\label{sec:llm_compare}
In addition to personalized text prompts obtained through image-to-text inversion using HiPer~\cite{han2023hiper}, we compare our approach to a different text prompt sampling technique using large language models (LLMs).
Given an original text prompt $y$, we employ ChatGPT~\cite{OpenAI2023GPT4TR} to generate $K$ prompts from $y$, enriching the description of the object.
For instance, starting with ``A high-quality photo of an ice cream sundae'' as the original prompt, we obtain an augmented prompt like ``A high-quality photo of an ice cream sundae with fresh berries and mint leaves''.
Subsequently, we utilize each of these prompts to condition the corresponding particle in the VSD loss.
The results, shown in \cref{fig:augment_text2}, indicate that while LLM-based augmented text prompts also lead to diverse 3D generations, their quality and diversity are inferior compared to our image-to-text inversion method.

\subsection{Extension to 3DGS}

To expedite training, our approach can be extended to 3D Gaussian Splatting (3DGS)~\cite{kerbl3Dgaussians}. Since the 3DGS method requires a point cloud to start the optimization, we utilize recent generative text-to-3D approaches such as 3DTopia~\cite{hong20243dtopia} and Shape-E~\cite{shape-e} to obtain the initial shapes for our training. Specifically, we conducted experiments on 3DGS using the same initial shape from 3DTopia and multiple initial shapes from Shape-E. As shown in Figure 9, our method can generate high-quality and diverse 3D renderings using the 3DGS backbone. The 3DGS representation, compared to Instant-NGP, significantly reduces the training time from approximately 7 hours to about 2 hours when using 6 particles.

\begin{figure}[t!]
  \centering
  \includegraphics[width=.8\linewidth]{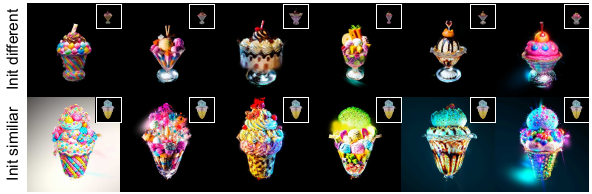}
   \caption{Our method also achieve remarkable diversity on 3DGS. We also show the initialized point-cloud at the top right corner of each sample.}\label{fig:3dgs}
\end{figure}

\section{Discussion and Conclusion}
\label{sec:conclusion}

\myheading{Limitations:}
Despite enhancing the diversity of the existing VSD framework, our model has limitations. It heavily relies on the HiPer inversion, which may struggle with outlier reference images, resulting in 3D models with unusual shapes and appearances. Also, our method shares VSD's limitations, such as the Janus problem, which can be addressed by orthogonal methods applicable to VSD.




\myheading{Conclusion:}
In this paper, we have successfully introduced a new text-to-3D synthesis method that focuses on diversifying the 3D generation by using 2D reference images and textual inversion to build augmented text prompts for conditioning the optimization. In future work, we plan to experiment with our augmented text embedding technique for other text-to-3D methods~\cite{lin2023magic3d,fantasia3d}, and more 3D representations~\cite{kerbl3Dgaussians}. 

\myheading{Acknowledgment.} Binh-Son Hua is supported by Science Foundation Ireland under the SFI Frontiers for the Future Programme (22/FFP-P/11522). 

\bibliographystyle{splncs04}
\bibliography{main}
\end{document}